\documentclass[letterpaper, 10 pt, conference]{ieeeconf}  

\IEEEoverridecommandlockouts                              
\usepackage{subfigure}
\usepackage{kantlipsum, lipsum}
\usepackage{booktabs}
\usepackage{bbding}
\usepackage{multirow}
\usepackage{graphicx}
\usepackage{microtype}
\usepackage{scalerel}
\usepackage{xspace}
\usepackage{bm}
\usepackage{bbm}
\usepackage{mathtools}
\usepackage{soul}
\usepackage{epsfig}
\usepackage{graphicx}
\usepackage{amssymb}
\usepackage{colortbl}
\usepackage{csquotes}
\usepackage{setspace}
\usepackage{adjustbox}
\usepackage{tabularx,ragged2e}
\usepackage{placeins}
\usepackage{amsmath}
\usepackage{tcolorbox}
\usepackage{xcolor}
\usepackage{url}
\usepackage{pifont}
\usepackage{blindtext}

\newcommand{\answerTODO}[1][]{\textcolor{red}{\bf [TODO]}}

\colorlet{darkgreen}{green!65!black}
\colorlet{darkred}{red!80!black}

\usepackage[symbol]{footmisc}
\usepackage{nameref}
\usepackage{varioref}
\usepackage[pagebackref=false,breaklinks=false,%
            colorlinks=true,bookmarks=true]{hyperref}
\usepackage[noabbrev,capitalize]{cleveref}
\usepackage{etoc}

\usepackage[utf8]{inputenc}
\usepackage[english]{babel}

\graphicspath{{figures/}}

\graphicspath{{figures/}}

\title{\LARGE \bf
Summit Vitals: Multi-Camera and Multi-Signal Biosensing at High Altitudes
}

\author{Ke Liu$^{1*}$, Jiankai Tang$^{2*}$, Zhang Jiang$^2$, Yuntao Wang$^{2\dagger}$, Xiaojing Liu$^{1\dagger}$, Dong Li$^1$, Yuanchun Shi$^{1,2}$ \\
\textbf{} 
\small{$^{*}$Co-first Author, $^{\dagger}$Co-corresponding Author, 1 Qinghai University, 2 Tsinghua University}}

\begin{document}

\maketitle
\thispagestyle{empty}
\pagestyle{empty}

\begin{abstract}

Video photoplethysmography (vPPG) is an emerging method for non-invasive and convenient measurement of physiological signals, utilizing two primary approaches: remote video PPG (rPPG) and contact video PPG (cPPG). Monitoring vitals in high-altitude environments, where heart rates tend to increase and blood oxygen levels often decrease, presents significant challenges. To address these issues, we introduce the SUMS dataset comprising 80 synchronized non-contact facial and contact finger videos from 10 subjects during exercise and oxygen recovery scenarios, capturing PPG, respiration rate (RR), and SpO2. This dataset is designed to validate video vitals estimation algorithms and compare facial rPPG with finger cPPG. Additionally, fusing videos from different positions (i.e., face and finger) reduces the mean absolute error (MAE) of SpO2 predictions by 7.6\% and 10.6\% compared to only face and only finger, respectively. In cross-subject evaluation, we achieve an MAE of less than 0.5 BPM for HR estimation and 2.5\% for SpO2 estimation, demonstrating the precision of our multi-camera fusion techniques. Our findings suggest that simultaneous training on multiple indicators, such as PPG and blood oxygen, can reduce MAE in SpO2 estimation by 17.8\%.

\end{abstract}

\section{INTRODUCTION}

Monitoring physiological signals in high-altitude environments presents unique challenges and opportunities due to the common increased heart rates and decreased blood oxygen levels, which lead to significant physiological fluctuations and potential discomfort~\cite{AKSEL2019121}. Traditional methods for measuring vital signs, such as heart rate (HR) and blood oxygen saturation (SpO2), rely on cumbersome and invasive devices, making them impractical for continuous and widespread use in these challenging environments. In response, remote photoplethysmography (rPPG) provides a non-invasive and convenient alternative, using cameras to detect subtle changes in light reflected from the skin, ideal for less intrusive monitoring in high-altitude environments \cite{mcduff2021camera}.

Research has shown that physiological parameters, such as SpO2, can vary significantly at high altitudes, thereby impacting the accuracy and reliability of health monitoring systems \cite{10151779}. Such environments necessitate advanced techniques and innovative methodologies for effective physiological monitoring. While public benchmark datasets are invaluable to the scientific community, most fail to capture the unique challenges of high-altitude environments. For instance, rPPG datasets like PURE~\cite{stricker2014non} and UBFC-rPPG~\cite{bobbia2019unsupervised} are primarily designed for low-altitude, resting-state conditions, which do not adequately reflect the physiological fluctuations seen in high-altitude settings.

To address these challenges, we introduce \textbf{Su}mmit Vitals: \textbf{M}ulti-Camera and \textbf{M}ulti-Signal Biosensing at High Altitudes (SUMS). Our study explores the utilization of the SUMS dataset with multi-camera systems to enhance physiological sensing in high-altitude conditions. This dataset includes physiological signals from 10 subjects captured in 80 videos, both non-contact facial videos and contact-based finger videos, synchronized at the millisecond level with accurate physiological data from oximeters and respiratory bands. These recordings cover various scenarios, such as exercise and oxygen recovery, to closely mirror the physiological changes that occur at high altitudes \cite{7893299,7985875}.

We assess the performance of both facial rPPG algorithms and finger contact PPG (cPPG) algorithms, demonstrating the enhanced accuracy of non-contact measurements through facial rPPG. Moreover, we investigate the potential of video fusion from different camera positions in predicting physiological indicators, achieving a quantified reduction with 7.6\% and 10.6\% reduction in mean absolute error (MAE) for SpO2 predictions. These methodological advancements significantly contribute to the field of remote health monitoring by providing a robust foundation for future research and development in non-invasive physiological measurement.

Our contributions are outlined as follows:
\begin{enumerate}
    \item We present the SUMS dataset, a comprehensive collection of synchronized non-contact and contact video data with true physiological readings, encapsulating varied scenarios indicative of high-altitude physiological challenges.
    \item We validate the accuracy of facial rPPG over finger cPPG algorithms, affirming the superiority of non-contact measurements for physiological monitoring.
    \item We demonstrate the effectiveness of multi-camera video fusion in improving the prediction accuracy of physiological indicators, underscored by significant reductions in prediction errors.
\end{enumerate}

    \begin{table*}[t!]
        \centering
        \caption{Dataset comparison}
        \begin{tabular}{ccccccc}
        \toprule 
        Dataset & Videos & Camera-Position & Vitals & Exercise & High-Altitude& Open-Source \\
        \midrule 
        \midrule 
        UBFC-rPPG~\cite{bobbia2019unsupervised} & 42 & Face & PPG &{\XSolidBrush}&\XSolidBrush &\Checkmark\\
        PURE~\cite{stricker2014non}&40& Face&PPG/SpO2&\XSolidBrush&\XSolidBrush&\Checkmark\\
        MMPD~\cite{tang2023mmpd}& 660 & Face  &PPG&\Checkmark&\XSolidBrush&\Checkmark\\
        SCOIHS~\cite{hoffman2021smartphonecameraoximetryinduced}& 6 & Finger  &PPG/SpO2&\XSolidBrush&Simulated&\Checkmark\\
        SSL~\cite{10151779}& 28 & Face &PPG/SpO2&\XSolidBrush&\Checkmark&\XSolidBrush\\ 
        
        NCTUBO~\cite{10151779}& 46 &Face&PPG/SpO2&\XSolidBrush &\XSolidBrush&\XSolidBrush\\ 

        \midrule
        SUMS(Ours) & 80 & Face+Finger & PPG/RR/SpO2 &\Checkmark &\Checkmark&\Checkmark\\
        \bottomrule 
        \end{tabular}\\
        \label{tab:datasets_compare}
        
    \end{table*}

\begin{figure}[htp]
    \centering
    \vspace{0.5cm}
    \includegraphics[width=0.47\textwidth]{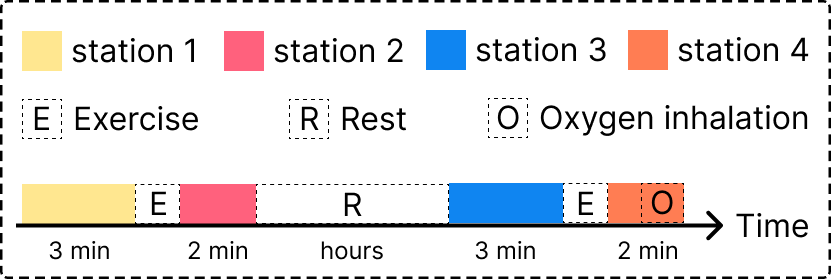}
    \vspace{0cm}
    \caption{A visual illustration of our data collection protocol. Participants have different activities (exercise, rest, oxygen inhalation) between stations (yellow, magenta, blue, and orange color).}
    \label{fig:Experiment_Process}
\end{figure}

\section{Related Works}

\subsection{Physiological Sensing in Plateau}
In recent years, numerous studies have focused on utilizing machine learning to predict physiological parameters such as HR, respiratory rate (RR), and SpO2 from photoplethysmography (PPG) signals~\cite{10017171,9856498,10075465}. Research shows that physiological parameters in low-altitude areas often remain within a narrow stable range, lacking data diversity. This issue is particularly pronounced for SpO2~\cite{Peacock1063,7893299}. Consequently, SpO2 prediction models tend to be inaccurate in high-altitude regions, where human SpO2 levels are generally lower than those at low altitudes~\cite{10151779}. To enhance data diversity, researchers have measured physiological parameters under conditions such as exercise, breath-holding, and low-pressure environments~\cite{s20174879,10.3389/fphys.2022.933397}. However, few studies have directly collected physiological data from individuals in high-altitude regions.

Despite this, the detection of physiological parameters like SpO2 in high-altitude areas has wide applications. In high altitudes areas hypoxemia is related to high-altitude illness~\cite{AKSEL2019121} and the incidence of acute mountain sickness (AMS) ranges from 15\% to 80\%, depending on ascent rate, absolute altitude reached, and individual susceptibility~\cite{doi:10.1089/ham.2010.1039}. AMS often leads to hypoxemia. Therefore, effective and user-friendly health monitoring models and warning systems are valuable for people in high-altitude regions.

Study~\cite{article} simulated high-altitude scenarios from 2000 meters to 4000 meters by creating low-pressure environment and collected heart rate, respiratory waveform, and SpO2 data from 15 subjects. Study~\cite{7893299} utilized a low oxygen partial pressure environment to simulate high altitude, causing subjects' SpO2 levels to range from 70\% to 100\% and a model was trained to predict SpO2 based on these physiological data.

Study~\cite{7985875} collected SpO2 data using a wrist-worn finger reflective pulse oximeter from subjects at altitudes of 5,360 meters and 4,259 meters. The aim was to derive a correlation coefficient R related to altitude as a correction parameter for high-altitude SpO2 measurements. However, this study had only three subjects, leading to inconclusive results. Study~\cite{10151779} collected facial PPG signals from subjects at an altitude of 3,150 meters using an RGB camera. They proposed a motion-resistant method using a KNN model to address motion artifacts, achieving an SpO2 prediction MAE of 4.39\%.
\subsection{Physiological Sensing Dataset}
There are a number of commonly used rPPG datasets (PURE~\cite{stricker2014non}, MAHNOB-HCI~\cite{soleymani2011multimodal}, BP4D~\cite{zhang2016multimodal}, VIPL-HR~\cite{niu2018vipl}, UBFC-Phys~\cite{sabour2021ubfc},   COHFACE~\cite{heusch2017reproducible}, UBFC-rPPG~\cite{bobbia2019unsupervised}, MMPD~\cite{tang2023mmpd}, and Scamps~\cite{mcduff2022scamps}) collected at different scenarios comprising facial videos and PPG signal, but none of them are focused on the high-altitude challenges.

Some datasets are either collected at high altitudes or simulate these conditions through breath-holding or oxygen-control techniques. However, only NCTUBO~\cite{hoffman2021smartphonecameraoximetryinduced} makes its dataset open-source, enabling further research. The SSL dataset~\cite{10151779}, recorded at an altitude of 3150 m in Taiwan, involved 14 subjects and featured two scenarios (static and dynamic SpO2) and two postures (lying down and sitting up). It used three cameras (FLIR Blackfly BFLY-U3-03S2C, 60 frames/s, Logitech C920, 30 frames/s, Samsung Galaxy A42 or Samsung Galaxy A60, 30 frames/s) and two pulse oximeters (An FDA-approved MASIMO MightySat fingertip pulse oximeter and a CMS50E+ oximeter) to ensure accurate SpO2 measurements and had a wide SpO2 distribution (mean: 87.98\%, SD: 5.15\%). The NCTUBO dataset~\cite{10151779}, recorded with the same devices at sea level, included 46 subjects and focused on the sitting posture with head support to minimize motion noise. It used a breath-holding method to reduce SpO2 levels, although only 1.21\% of instances had SpO2 below 90\%, indicating data imbalance. We picked several most related datasets for analysis and comparison as shown in Table \ref{tab:datasets_compare}.

\subsection{Physiological Sensing Model Training}
To improve the prediction of various physiological parameters, our study simultaneously collects facial and finger videos from participants. We use a single model to predict heart rate, respiratory rate, and SpO2 simultaneously, thereby achieving better training performance. Similar approaches have been used in previous studies to enhance prediction accuracy for multiple physiological parameters. Study~\cite{pmlr-v184-ren22a} designed a loss function based on the correlation between respiratory rate and heart rate to improve the accuracy of both predictions. Study~\cite{9903413} employed the SMP-Net model to process multimodal videos in visible and infrared light to predict heart rate and respiratory rate simultaneously. Study~\cite{9082808} used the PP-Net model to predict heart rate, systolic blood pressure, and diastolic blood pressure simultaneously, demonstrating good performance in a large and diverse population. Study~\cite{9747109} employed MultiPhys to impute missing heart rate or SpO2 data in open datasets and used machine learning to predict both parameters from facial videos, showing improved prediction performance.

   \begin{figure}[htp]
    \centering
    \includegraphics[width=0.4\textwidth]{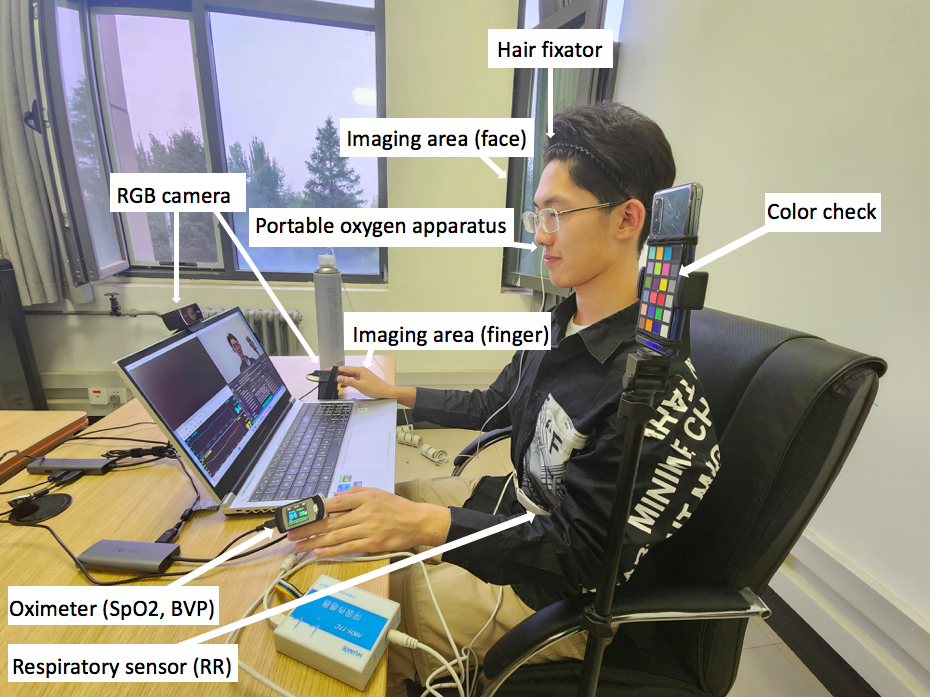}

    \caption{The experimental setup of data collection while participants are having oxygen inhalation.
    }
    \vspace{-0.5cm}
    \label{fig:device}
    \end{figure}

\section{Dataset}

Approved by the Institutional Review Board, we recruited ten participants living on the Qinghai Plateau(2274m-4898m) to collect video and physiological data in a real high-altitude environment. Data collection utilized two Logitech C922 cameras to capture videos of participants' faces and fingers, along with a CMS50E+ oximeter for measuring PPG signals and blood oxygen levels. Respiratory patterns were monitored using an HKH-11C respiratory sensor. Video recordings were captured at a resolution of 1280x720 pixels, running at 60 frames per second. The PPG signals were recorded at a frequency of 20 Hz, while respiratory waves were captured at 50 Hz. The collection platform was designed to capture all the frames and signals in millisecond synchronization based on the PhysRecorder\cite{wang2023physbench} framework. SUMS dataset is available at \url{https://github.com/thuhci/SUMS}. This section provides a detailed introduction to the dataset collection process, data processing methods, and the distribution of physiological signals in the dataset.

\subsection{Data Collection} 
    \begin{figure*}[htbp]

    \centering
        \subfigure[BVP]{
        \includegraphics[width=0.45\textwidth]{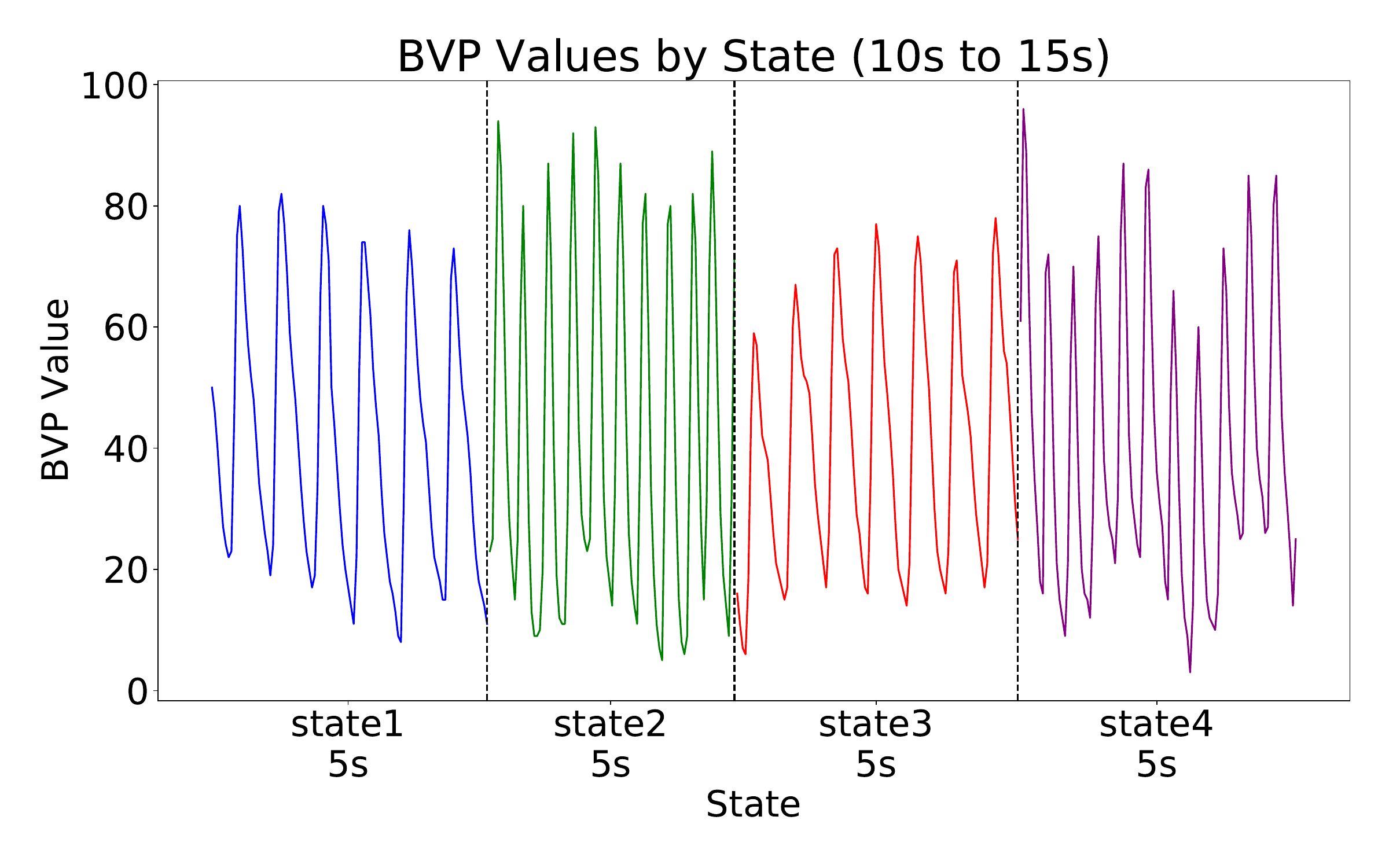}
    }
    \quad
    \centering
        \subfigure[SpO2]{
        \includegraphics[width=0.45\textwidth]{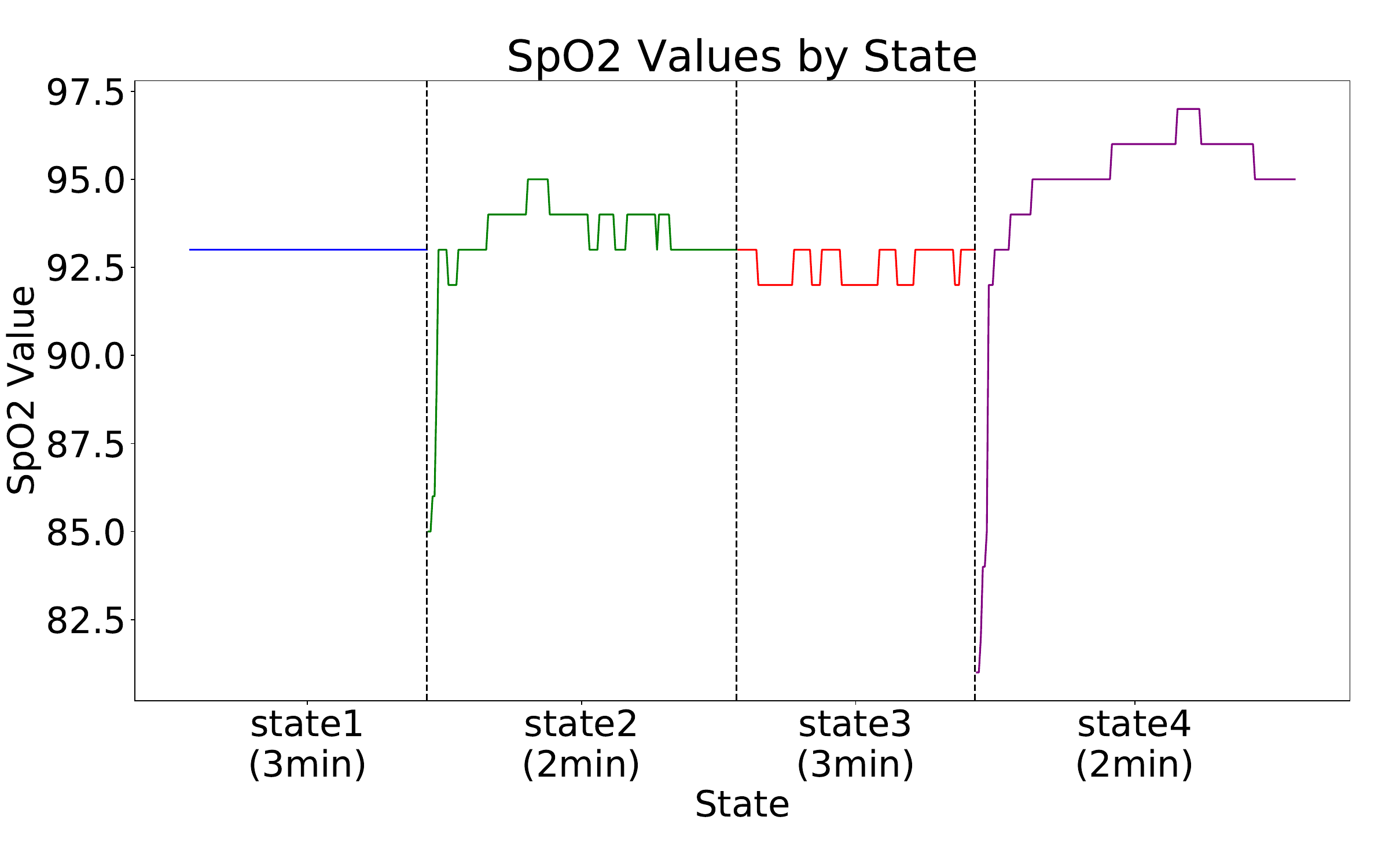}
    }
    \quad
    \centering
        \subfigure[HR]{
        \includegraphics[width=0.45\textwidth]{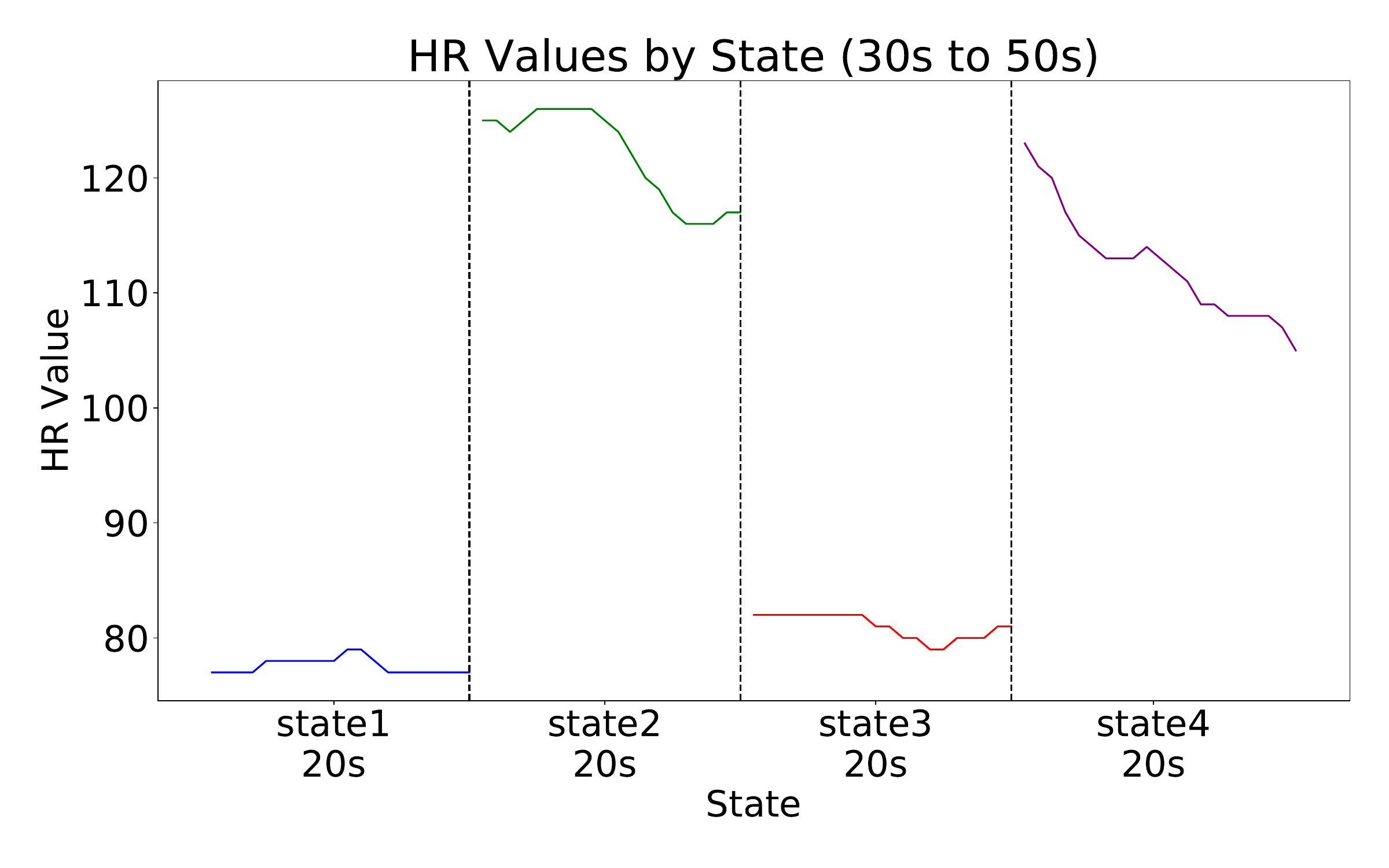}
    }
    \quad
    \centering
        \subfigure[RR]{
        \includegraphics[width=0.45\textwidth]{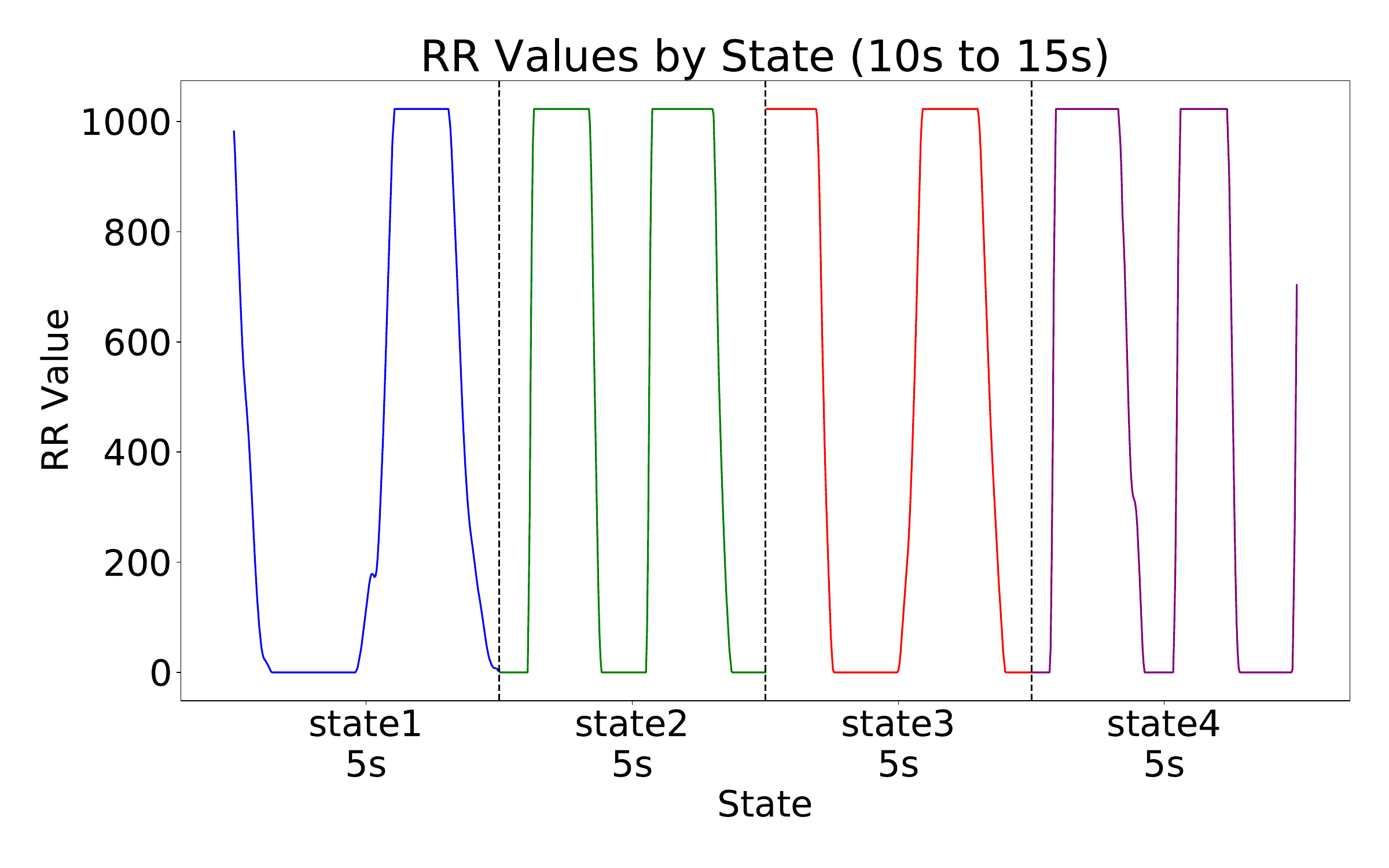}
    }

    \caption{
    \textbf{Physiological indicators of the same subject under four states.} 
    (a) BVP (b) SpO2 (c) HR (d) RR. The duration of the signal is noted in the bottom of the figure.}
    \label{fig:Physiological_indicators}
    \end{figure*}

    \begin{figure*}[htp]
            \centering
        \subfigure[HR distribution]{
        \includegraphics[width=0.45\textwidth]{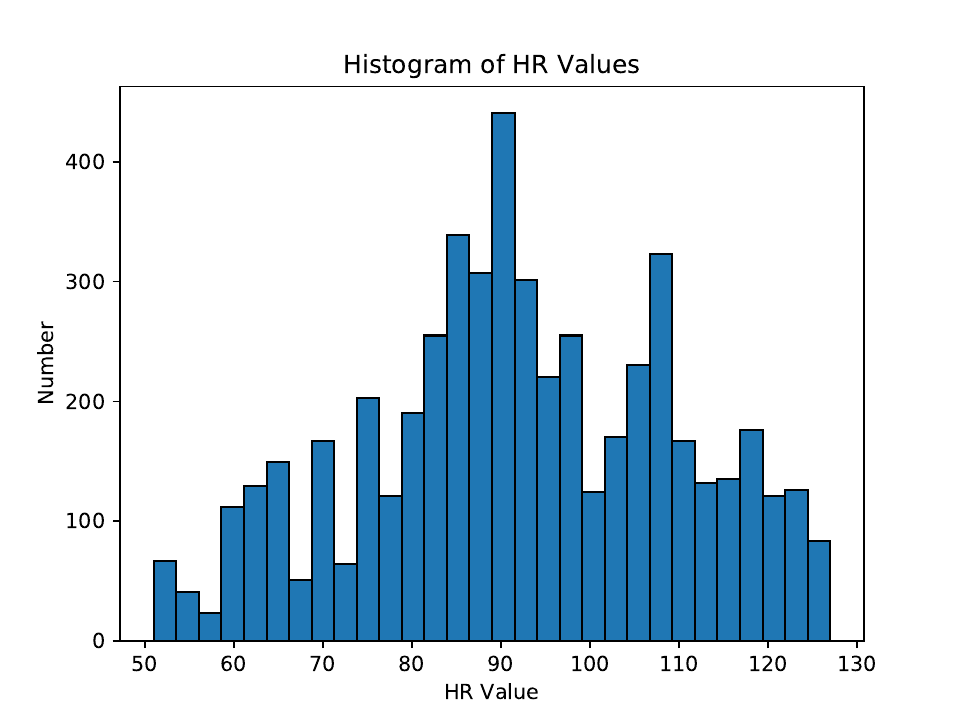}
        }
    \quad
    \centering
        \subfigure[SpO2 distribution]{
        \includegraphics[width=0.45\textwidth]{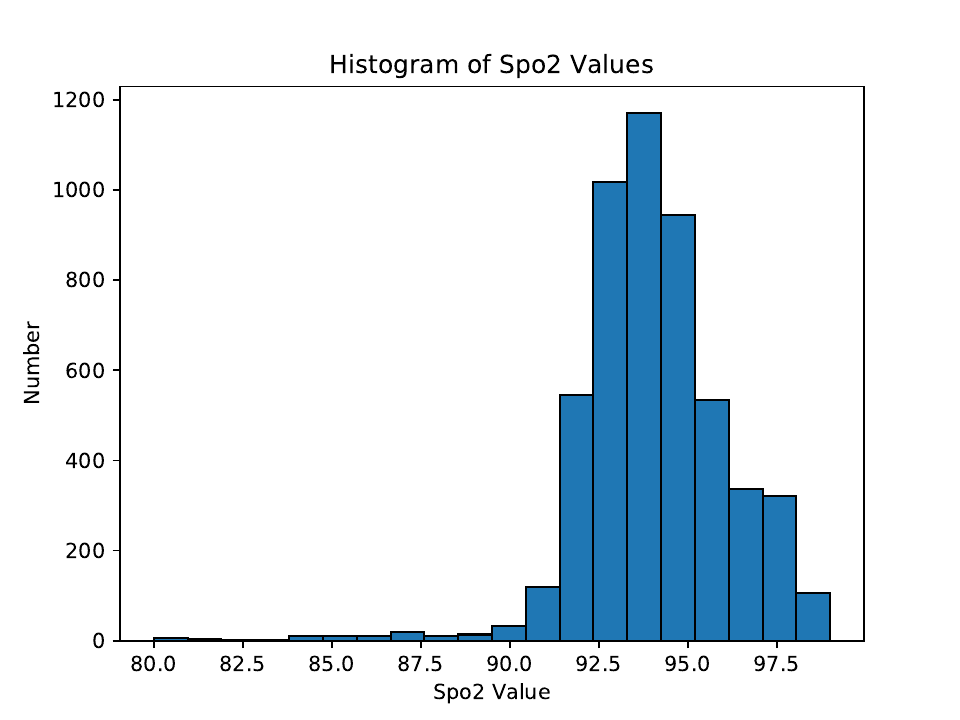}
        }
    \caption{
    \textbf{HR and SPO2 distribution statistics of all samples.} }
    \label{fig:hr_spo2}
    \end{figure*}

The experimental process is depicted in Figure \ref{fig:Experiment_Process}. We designed two sets of experiments, each consisting of a rest state and a recovery state. The first set of experiments was conducted in the morning. During the rest state (state 1), subjects wore the HKH-11C breathing sensor on their abdomen, used the CMS50E+ oximeter on their left index finger, and positioned their right index finger on the Logitech C922 camera. They were instructed to keep their heads still and gaze at a screen, while data was recorded for three minutes. Following the rest period, the subjects engaged in a physical activity where they quickly climbed stairs. Post-exercise, they retained the same equipment setup as in state 1 and data was recorded for two minutes, marking this as state 2.

The second set of experiments occurred in the afternoon, to ensure the subjects recovered completely. In this set, the subjects again began in a resting state, akin to state 1, for three minutes, designated as state 3. After resting adequately, they climbed the stairs rapidly. Unlike the first experiment, during the post-exercise recording, subjects wore portable oxygen devices while maintaining the same equipment setup as in state 1. This post-exercise recording lasted two minutes and was labeled state 4.

The experimental protocol identified four distinct states across the experiments: state 1, state 2, state 3, and state 4. The physical exertion aimed to decrease blood oxygen saturation levels, while the introduction of supplemental oxygen was intended to amplify the fluctuations in blood oxygen saturation.

Post-exercise observations indicated that most subjects' blood oxygen levels dropped below 92\%, although a few maintained higher levels, likely due to regular exercise habits that contribute to more stable oxygen saturation. Subjects were given ample rest to recover before proceeding to the next experimental phase. Figure \ref{fig:device} illustrates the setup used during these experiments.

\subsection{Data Processing}
In our study, critical preprocessing steps were applied to enhance the rPPG signal detection from both face and finger videos. Facial video analysis involved detection, tracking, and differential processing to extract robust periodic signals, crucial for accurate rPPG tasks. Similarly, differential techniques were also applied to the videos of the fingers, enhancing the signal clarity of physiological measures. Figure \ref{fig:Network_Model}
illustrates these preprocessing steps. To align the PPG signals from the CMS50E+ oximeter with the video frame rate, the signals were resampled from 20 Hz to 60 Hz. Using the modified PhysRecorder framework, we ensured that timestamps from all recording devices were accurately synchronized and stored in CSV format, facilitating precise alignment with the video timestamps.

    
\subsection{Physiological Signal Distribution}
Figure \ref{fig:Physiological_indicators}  and Figure \ref{fig:hr_spo2} display the variations in blood volume pulse (BVP), HR, RR and SpO2 across different experimental states. Notably, several instances of hypoxemia were recorded, with SpO2 levels dropping below 92\% during physical exertion. These findings underscore the potential of our dataset to represent medical and rehabilitative scenarios, especially given the inclusion of oxygen therapy, distinguishing our work from other datasets and demonstrating the practical challenges of deploying rPPG technologies in real-world conditions.

\section{Method}
\subsection{Input embedding}

    \begin{figure*}[htp]
    \centering
    \vspace{0.5cm}
    \includegraphics[width=1\textwidth]{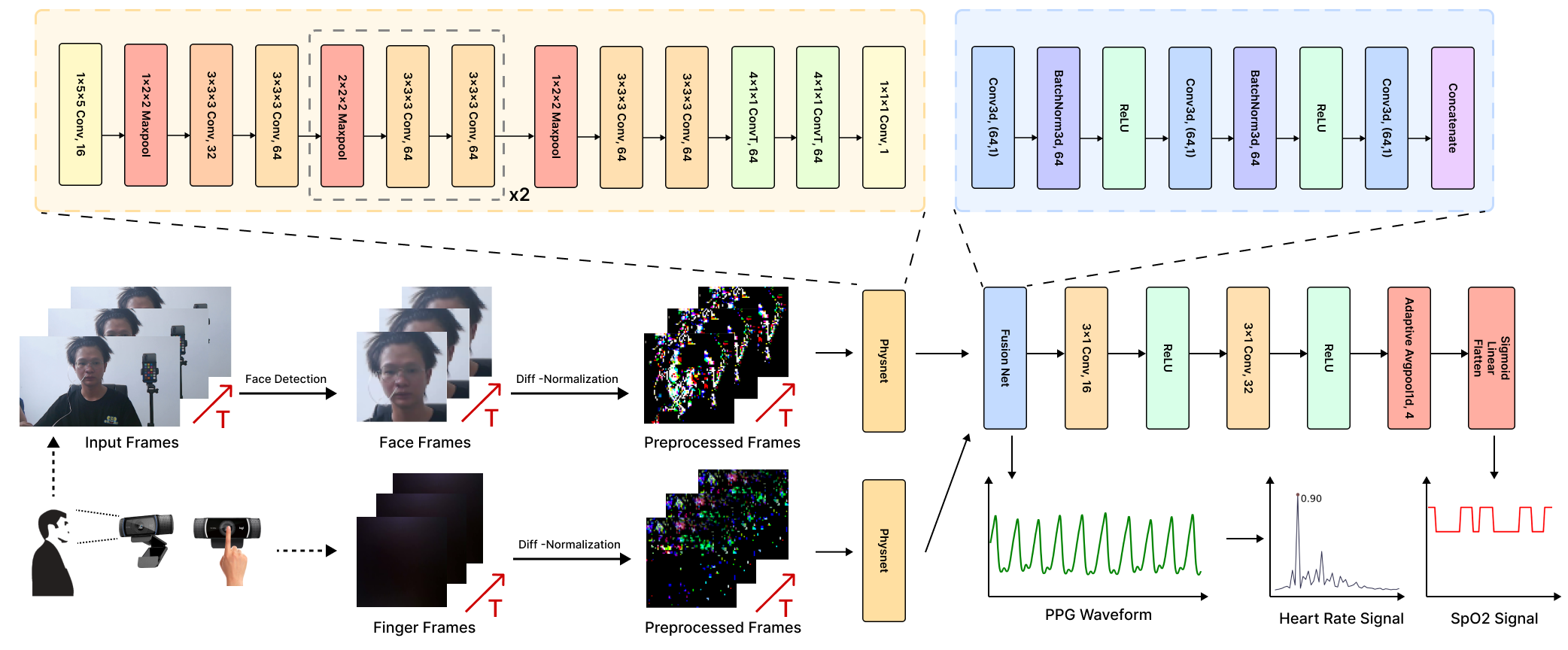}
    \vspace{0cm}
    \caption{\textbf{MultiPhysNet Model with Input frames of face and finger.} PPG and SpO2 estimation tasks are trained simultaneously with combined loss.}
    \label{fig:Network_Model}
    \end{figure*}

This paper highlights an innovation in input embedding, demonstrated by the model’s ability to process video signals from two distinct sources: facial video and finger video. As illustrated in Figure \ref{fig:Network_Model}, the model first preprocesses the facial and finger video data, converting them into frame sequences. These frame sequences are then feature-extracted using the same 3D convolution encoder and combined through the fusion module (FusionNet).

This design enables the model to simultaneously utilize information from both facial and finger videos. Signals from different body parts provide complementary information, enhancing the model’s adaptability to diverse inputs. Facial videos primarily reflect changes in facial blood flow, while finger videos offer more direct blood pulsation signals. By fusing these signals, the model captures changes in physiological characteristics more comprehensively. This innovative design enhances the model’s performance in practical applications and provides a more accurate and effective foundation for subsequent heart rate and blood oxygen saturation prediction.

\subsection{Neural Network Model}
\begin{table*}[htbp]
    \small
	\caption{Intra-Dataset and Cross-Dataset Remote HR Estimation Results.}
	\vspace{0.3cm}
	\label{tab:cross_dataset_results}
	\centering
	\small
	\setlength\tabcolsep{3pt} 
	\begin{tabular}{r|ccc|ccc|ccc|ccc}
	\toprule

	    \textbf{Testing Set} & \multicolumn{3}{c}{Ours} & \multicolumn{3}{c}{\textbf{PURE} \cite{stricker2014non}} & \multicolumn{3}{c}{\textbf{UBFC-rPPG} \cite{sabour2021ubfc}} & \multicolumn{3}{c}{\textbf{MMPD\_S} \cite{tang2023mmpd}} \\
        \textbf{Training Set}& MAE$\downarrow$ & MAPE$\downarrow$ & $\rho$ $\uparrow$ & MAE$\downarrow$ & MAPE$\downarrow$ & $\rho$ $\uparrow$ & MAE$\downarrow$ & MAPE$\downarrow$ & $\rho$ $\uparrow$ & MAE$\downarrow$ & MAPE$\downarrow$ & $\rho$ $\uparrow$ \\ \hline  \hline 
        Ours  & 0.30 & 0.37 & 1.00 & \textbf{4.64} & \textbf{4.34} & \textbf{0.75} & 3.81 & 3.46 & 0.79 & 6.16 & 8.23 & 0.45 \\
        \textbf{PURE} \cite{stricker2014non}& 5.82 & 6.23 & 0.75 & 4.17 & 4.65 & 0.76 & 1.84 & 1.88 & 0.95 & 4.23 & 5.48 & 0.70 \\
        \textbf{UBFC-rPPG} \cite{sabour2021ubfc} & 9.76 & 12.80 & 0.40 & 6.73 & 11.04 & 0.68 & 0.10& 0.08 & 1.00 & 4.04 & 6.11 & 0.74 \\
        \textbf{MMPD\_S} \cite{tang2023mmpd}& 12.79 & 16.12 & 0.48 & 15.18 & 25.74 & 0.38 & 2.34 & 2.48 & 0.95 & 1.10 & 1.27 & 0.98 \\
       
       \bottomrule 
       \end{tabular}
       \\
       \footnotesize
      MAE = Mean Absolute Error in HR estimation (Beats/Min), MAPE = Mean Percentage Error (\%), $\rho$ = Pearson Correlation in PPG Prediction.
\label{tab:intra_cross_dataset}
\end{table*}

The overall architecture of our MultiPhysNet is shown in Figure \ref{fig:Network_Model}. Our MultiPhysNet incorporates a FusionNet to effectively integrate features from both facial and finger videos. This design leverages their complementary information by combining signals from different body parts. Facial video provides features related to facial blood flow, whereas finger video captures pulsation signals. Combining these signals helps mitigate the noise that might affect individual signals, thereby enhancing the accuracy of heart rate and blood oxygen saturation predictions.

Additionally, building on the original HR prediction, our MultiPhysNet now includes a SpO2 prediction module. This enhancement allows the model to predict both heart rate and blood oxygen saturation. By integrating information from both facial and finger videos, this design enhances the comprehensive understanding of these two key physiological indicators and improves the model's practicality and accuracy.

\subsection{Joint-Training}

We propose an enhanced joint training mechanism that simultaneously optimizes HR and SpO2 prediction targets, significantly improving overall performance. Unlike traditional single-task training methods, our approach involves a novel loss function specifically crafted to handle both HR and SpO2 tasks simultaneously. By jointly optimizing these two objectives during training, the model learns a more comprehensive feature representation, which enhances its ability to predict complex physiological signals. 

The custom loss function we designed is aimed at improving the precision of physiological signal estimation. It is formulated as:

\small
\[
\text{Loss} = \text{MSE}_{\text{BVP}} \times 100 + \text{MSE}_{\text{SpO}_2} \times (100 - \text{Mean}(\text{SpO}_2))
\]
\normalsize

where $\text{MSE}_{\text{BVP}}$ represents the mean squared error of the blood volume pulse (BVP) signal, and $\text{MSE}_{\text{SpO}_2}$ represents the mean squared error of the SpO$_2$ signal. The factor $(100 - \text{Mean}(\text{SpO}_2))$ emphasizes the importance of accurate SpO$_2$ measurements, particularly under hypoxic conditions, where SpO$_2$ levels are lower. This formulation aims to balance the precision of both BVP and SpO$_2$ measurements, leading to a more robust model for multimodal physiological signal measurement.

\section{Result}

All experiments were conducted using the improved training framework of rPPG-Toolbox\cite{liu2024rppg}. The environment configuration included PyTorch 2.2.2+cuda12.1, Python 3.8, and NVIDIA GeForce RTX 3090.
\subsection{Remote Heart Rate Sensing}
We experimented on four datasets using the improved training framework of rppg-toolbox\cite{liu2024rppg}, with a learning rate of 9e-3, 30 epochs, and a batch size of 16.

In this paper, we studied how PhysNet\cite{wang2023physbench} performs on our datasets. For intra-dataset experiments, 80\% of the data was used as the training set and 20\% as the test set, with no validation set. For cross-dataset experiments, 80\% of the data was used as the training set, 20\% as the validation set, and all corresponding datasets, except MMPD\cite{tang2023mmpd}, were used as the test set. We selected only a subset of MMPD\cite{tang2023mmpd}, which contains 8 labels. For training, validation, and testing, we selected the conditions LED-low, LED-high, Stationary, and Talking, and only included skin color 3, thus forming the subset MMPD\_S\cite{tang2023mmpd}.

Table \ref{tab:cross_dataset_results} presents the results of the supervised neural network in both intra-dataset and cross-dataset evaluations. The results indicate that by switching the training set from the UBFC-rPPG dataset to our dataset, the MAE on the PURE dataset decreased from 6.73 to 4.64. This represents an error reduction of approximately 31.06\%, demonstrating the effectiveness of our dataset in improving model performance. We found that using our dataset as the training set and the PURE dataset as the test set yielded better results than the reverse. This trend is observed not only on the PURE dataset but also on other datasets. This suggests that the data collected in high-altitude environments and during exercise is more diverse, as it includes both still videos and recovery videos with elevated heart rates post-exercise.

\subsection{HR-SpO2 Multi-task Learning}
We split the dataset into 80\% for training and validation and 20\% for testing. Within the 80\% allocated for training and validation, we used states 1, 3, and 4 for training and state 2 for validation. We established different training tasks, including individual training for HR, individual training for SpO2, and joint training for HR-SpO2. 

Table \ref{tab:Multi-task} presents the results of the different training tasks. The results indicated that in facial signal analysis, jointly trained HR-SpO2 performed better than separately trained SpO2, while ensuring HR stability. Specifically, the MAE decreased from 3.20 to 2.63, resulting in a 17.8\% reduction in error. This suggests that combined training can enhance the performance of the blood oxygen task.
\subsection{Multi-Camera Sensing}
In this experiment, we used two Logitech C922 cameras of the same type: one for facial video and one for finger video. With the multi-camera setup, we aim to improve the accuracy and reliability of remote physiological signal measurement by combining data from different parts of the body.

Table \ref{tab:Multi-task} presents the results of multi-camera fusion for various tasks. The fused multiple-camera system achieved the best results in SpO2 prediction for the joint HR-SpO2 task. Specifically, the error was reduced by 7.6\% compared to using facial signals alone, and by 10.6\% compared to using fingertip signals alone. This demonstrates that multi-camera fusion significantly enhances SpO2 prediction in joint HR-SpO2 tasks.

\begin{table*}[htbp]
\caption{The results of HR-SpO2 Multi-Task Training}
    \label{tab:Multi-Task Training}
    \centering
    \begin{tabular}{c|cc|cc|cc|cc}
    \toprule
         &  \multicolumn{2}{c}{\textbf{Single HR Task}} &  \multicolumn{2}{c}{\textbf{Single SpO2 Task}}&  \multicolumn{2}{c}{\textbf{Both HR Task}}&  \multicolumn{2}{c}{\textbf{Both SpO2 Task}}\\
          \textbf{Training Set}& MAE$\downarrow$ & MAPE$\downarrow$ &MAE$\downarrow$ & MAPE$\downarrow$ & MAE$\downarrow$ & MAPE$\downarrow$ & MAE$\downarrow$ & MAPE$\downarrow$ 
         \\ \hline  \hline 
        Face-Camera & \underline{0.30}  & \underline{0.37} & 3.20 & 3.29  & \underline{0.30} & \underline{0.37} & 2.63 & 2.69   \\
        Finger-Camera & 4.12 & 4.28 & \textbf{2.33} & \textbf{2.38}  & 3.52 & 3.80 & 2.72 & 2.78   \\
        Both-Cameras & \textbf{0.05} & \textbf{0.07} & 2.81 & 2.88  & 3.13 & 3.28
         & \underline{2.43} & \underline{2.50}   \\
                \bottomrule 
    \end{tabular}\\
    \footnotesize
      MAE = Mean Absolute Error in HR estimation (Beats/Min), MAPE = Mean Percentage Error (\%). \textbf{Bold} means the best performance while \underline{underline} means the second best performance.
      \label{tab:Multi-task}
\end{table*}

\section{Discussion}

\subsection{Challenges in High Altitude}
High-altitude environments introduce two significant challenges in the realm of video-based physiological perception. First, physiological signals tend to exhibit increased variability, including elevated heart rates and reduced blood oxygen levels. These changes necessitate enhanced model robustness and adaptability. Current methodologies, trained on existing datasets, struggle to achieve a Mean Absolute Error (MAE) below 5 in our specialized SUMS dataset, as demonstrated in Table \ref{tab:cross_dataset_results}. Secondly, high-altitude conditions may trigger atypical health states such as hypoxemia (blood oxygen levels below 92\%), which can lead to cardiovascular system anomalies like altered blood flow rates. These abnormalities potentially undermine the efficacy of predictive models designed for stable conditions.
\subsection{Contact or Non-contact?}
Prior research typically favors contact-based methods for physiological signal measurement due to their perceived robustness. However, our experiments reveal that in relatively static conditions, remote video-based HR estimations outperform those obtained via contact methods using fingertip videos. This superiority may stem from the larger effective signal region provided by facial videos, which are less susceptible to minor disturbances that can disproportionately affect the smaller surface area of a fingertip. Despite this, fingertip videos remain superior in SpO2 measurement tasks. This observation prompts further contemplation about the future directions of video-based physiological sensing, particularly in how signal capture methods might evolve to enhance accuracy and reliability across different physiological parameters.
\subsection{Limitations}
While our study provides valuable insights into the capabilities of multi-camera systems in high-altitude settings, it has several limitations. The sample size of ten subjects might not fully represent the wide range of physiological responses to high altitudes across different demographics. Additionally, the video recordings were limited to specific controlled motions, which might not adequately simulate the more dynamic movements experienced in real-world high-altitude activities. Future research should aim to include a broader participant base and more varied motion scenarios to enhance the generalizability of the findings.


\section{CONCLUSIONS}

This study enhances the field of remote physiological monitoring at high altitudes using video photoplethysmography (vPPG) by distinguishing between non-contact facial rPPG and contact fingertip cPPG. Our focused contributions are:

\begin{enumerate}
    \item We introduced the SUMS dataset, the first of its kind to the best our knowledge, comprising 80 videos from 10 subjects, designed to improve the representation of physiological data in high-altitude settings, particularly under varied conditions such as exercise and oxygen recovery.
    \item Joint training on HR and SpO2 demonstrated a significant decrease in prediction error, reducing the MAE for SpO2 predictions by 17.8\% when compared to training on single task.
    \item Implementing multi-camera fusion techniques led to a further reduction in the MAE for SpO2, by 7.6\% using facial videos and 10.6\% using fingertip videos.
\end{enumerate}

These results underscore the benefits of combining multi-camera systems and joint predictive modeling to improve the accuracy and reliability of physiological monitoring systems in challenging environments. This research provides a valuable foundation for future advancements in non-invasive health monitoring technology.

\addtolength{\textheight}{-12cm}   





\section*{ACKNOWLEDGMENT}
This work is supported by the Natural Science Foundation of China (NSFC) under Grant No. 62366043, the Subtopic of Qinghai Top Ten Innovation Platform
Project (No.ZYYSDPT-2023-02),  Tsinghua University Initiative Scientific Research Program, Beijing Natural Science Foundation(No.QY23124), Beijing Key Lab of Networked Multimedia, and Institute for Artificial Intelligence, Tsinghua University. The experimental procedures involving human subjects described in this paper were approved by the Institutional Review Board. 



\bibliographystyle{unsrt}
\bibliography{MMPD}

\end{document}